\documentclass[10pt,conference,compsocconf]{IEEEtran}

\usepackage{url}
\usepackage{color, colortbl}
\definecolor{Gray}{gray}{0.9}
\usepackage{graphicx}	

\begin{document}
\title{Dynamic Gesture Recognition\protect\\ ETH Zurich}

\author{
  Costanza Maria Improta, Jonas Bokstaller\\
  cimprota@ethz.ch, bojonas@ethz.ch}

\maketitle

\begin{abstract}
The Human-Machine Interaction (HMI) research field is an important topic in machine learning that has been deeply investigated thanks to the rise of computing power in the last years. The first time, it is possible to use machine learning to classify images and/or videos instead of the traditional computer vision algorithms. The aim of this paper is to build a symbiosis between a convolutional neural network (CNN) \cite{cnn} and a recurrent neural network (RNN) \cite{rnn} to recognize cultural/anthropological Italian sign language gestures from videos. The CNN extracts important features that later are used by the RNN. With RNNs we are able to store temporal information inside the model to provide contextual information from previous frames to enhance the prediction accuracy. Our novel approach uses different data augmentation techniques and regularization methods from only RGB frames to avoid overfitting and provide a small generalization error.
\end{abstract}

\section{Introduction}
The task of video classification consists in choosing the right category out of 20 Italian sign types. This is a fundamental task as it can help deaf people in expressing their needs and socializing more easily by translating their sign language in real-time and by enabling them to have a conversation without barriers.

Over the last few years, computational infrastructure has become better than ever and this lead to breakthroughs in machine learning, with more and more research in the field focusing on CNNs and RNNs. These networks allow us to process efficiently and in real-time the entire amount of information of a whole video - an achievement that was thought to be impossible in the past. We make use of a particular type of CNN, namely a 3D-CNN \cite{3dcnn}, which is able to extract spatio-temporal features and to split, downscale and upscale each frame while efficiently applying different filters to it. With this transformed and filtered information, we are now able to feed it into the RNN, which models longer temporal relations throughout the video to finally classify the video into one of 20 categories by using the softmax activation function.

As training data we were given short video clips of humans performing different Italian sign language movements, which are categorized into 20 classes. The data was provided as a customized version of the ChaLearn dataset \cite{ChaLearn}.
\begin{figure}[h]
\includegraphics[width=\linewidth]{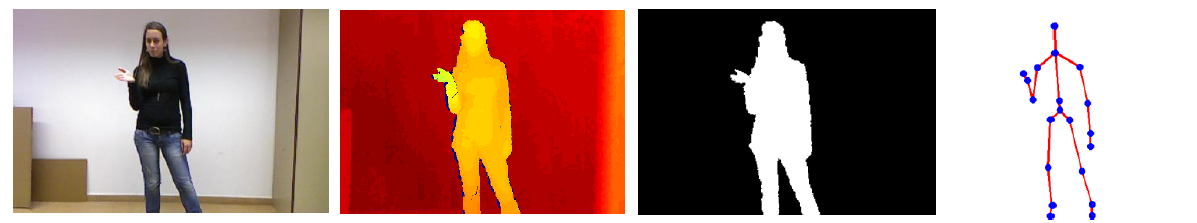}
\caption{Training data that was provided as a customized version of ChaLearn \cite{ChaLearn} dataset. Each sample contained RGB, depth, user segmentation and skeleton information.}
\end{figure}

\begin{figure}[h]
\includegraphics[width=\linewidth]{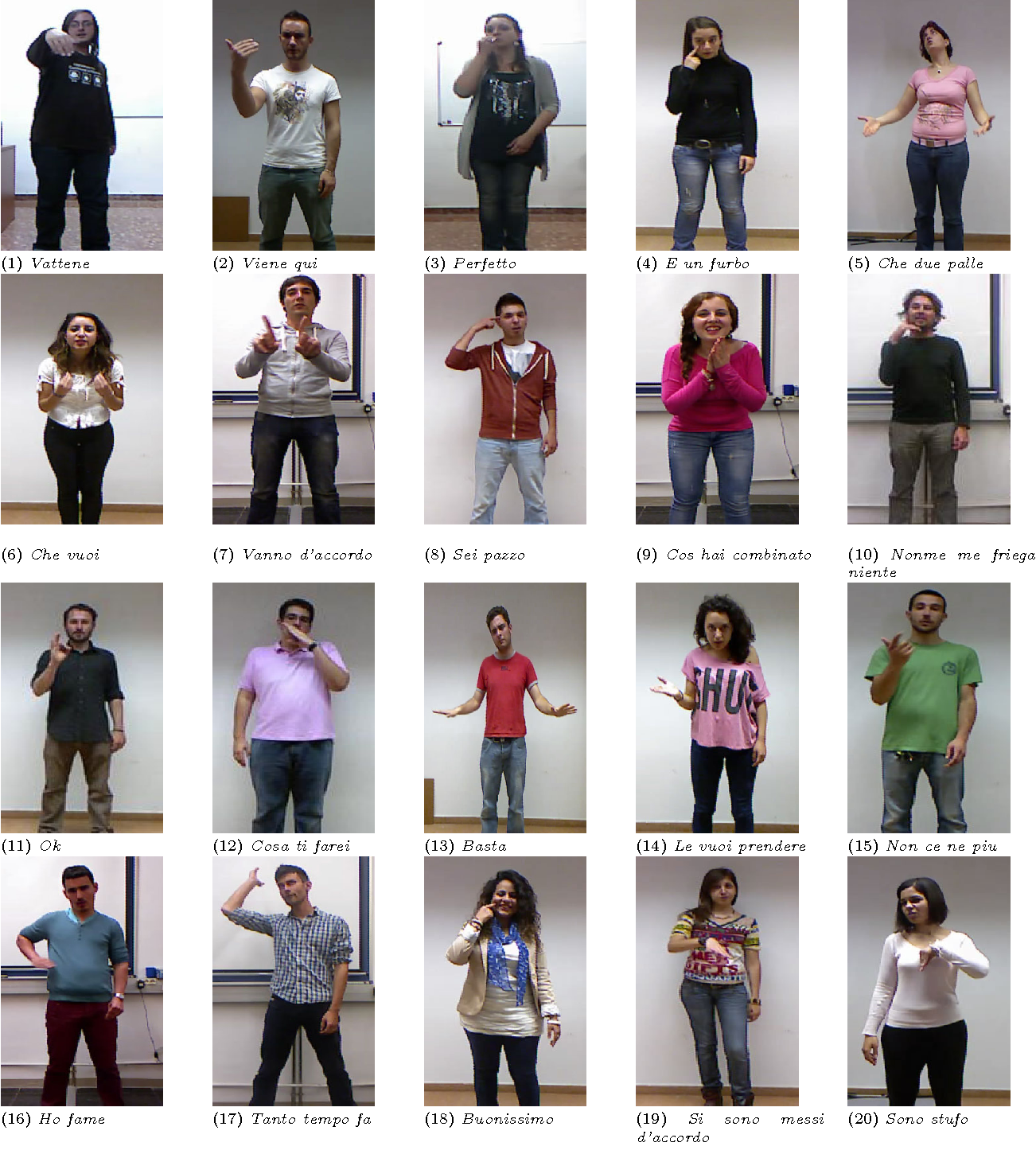}
\caption{Sample from each of the 20 gesture categories from the dataset. Each type of gesture represents a short phrase like "Sei pazzo" which translates to "You are crazy". \cite{mosaic}}
\end{figure}

The data is stored in the TFRecord format, which enables us to efficiently read the data linearly. All gestures were captures using a Microsoft Kinect sensor, and as such, the data holds multimodal information. Each data sample then contains RGB, depth information, segmentation mask, body skeleton, number of frames and label. Each sample is around 50-150 frames.

This paper gives the reader a brief overview over the different methods we implemented, the main challenges we faced, such as avoiding overfitting by using data augmentation, and the results we got by predicting the labels for the testing data.

\section{Related Work}
The dynamic gesture recognition field is vastly researched in the last couple of years. One of the main problems in this field is to recognize the gesture performed, because these can depend not only on the sequence, but also on the cultural aspect like we see in the Italian version of sign language.

Clebeson et al. \cite{multimod} use only RGB information because this is more interesting since RGB cameras are basically everywhere now. They call their technique to represent spatio-temporal information star RGB. For this, they use two Resnet CNNs, a soft-attention ensemble and a fully connected layer which classifies the gesture. They trained on the same dataset as we do and achieved an accuracy of around 94\%.

Kenneth et al. \cite{depthOnly} use another approach by only relying on depth and skeleton information for dynamic hand gesture recognition. They found out, that RNN perform well with only the skeleton information given. They use depth data, apply it to the CNN and extract important spatial information from the depth images. With both, RNN and CNN in place, they are capable of recognizing sequences of gestures more accurately.

\section{Model and Methods}
\subsection{General Workflow}
After a careful evaluation of the dataset and a thorough research of the state-of-the-art models for the given problem, we settled for the Conv3DRNN approach, and produced a vanilla implementation of it. The biggest challenge we encountered is overfitting, and we then devised some data augmentation methods and tested different logic loss functions to generalise the model.

\subsection{Data Augmentation}
No preprocessing has been carried out on the data. However, different data augmentation methods were devised to prevent overfitting, starting from the simplest one which involved the insertion of a small amount of Gaussian noise on the data. Concerning other spatial data augmentation methods, flipping, cropping, rotating, rescaling and adjusting hue and brightness were implemented. On the other hand, methods involving the reversal of the frame ordering, or the shortening of the length of the clip, were tackled to perform temporal data augmentation.

\subsection{Network Architecture}
As anticipated in the previous sections, our network comprises of a 3D-CNN network and an RNN network. Being the single RGB model network our best performing one, in this section we will focus only on its implementation, despite having also implemented a multimodal version of it.

\begin{figure}[h]
\includegraphics[width=\linewidth]{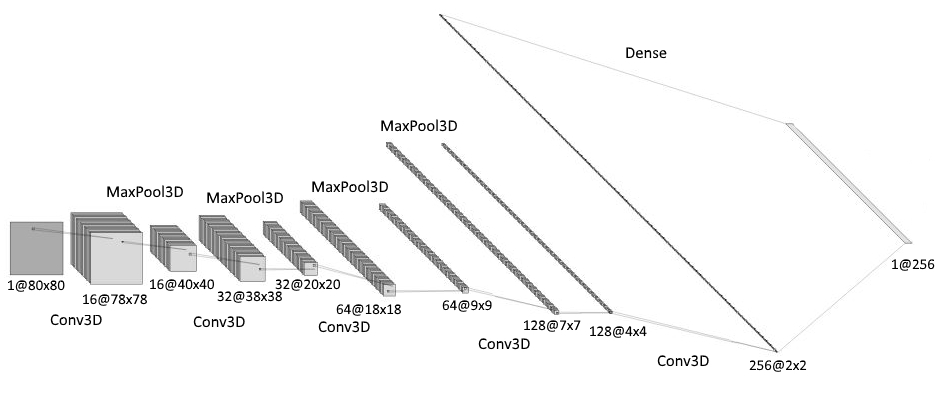}
\caption{Visualization of the CNN architecture we used. Five times the sequence of Conv3D, BatchNormalization and MaxPool3D have been repeated. Due to simplicity of the visualization, BatchNormalization has been left out. After that, flatten layer, dense and BatchNormalization followed to reduce the input to a vector of size 256 for the RNN.}
\end{figure}

The 3D-CNN for the RGB mode uses a Conv3D filter, making use of 3 filters and a kernel size of [3,3,3] with padding and a swish activation function. Batch normalization, followed by a 3D max pooling, is then applied to the output of the Conv3D. The input is then flattened and inputted to a dense layer with 256 hidden units, before applying batch normalization again.

\begin{figure}[h]
\centering \includegraphics[width=0.7\linewidth]{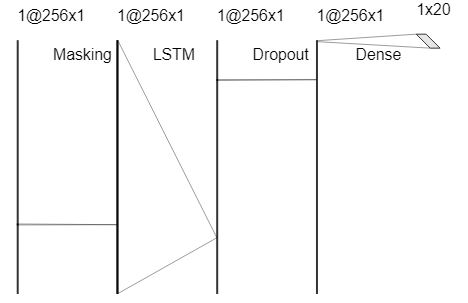}
\caption{Visualization of the RNN architecture we used. After inputting the vector from the CNN model, we apply masking layer, LSTM with 256 cells, dropout layer and finally end with a dense layer of 20 probabilities for each class.}
\end{figure}

The output of the 3D-CNN is then fed into the RNN part of the network, which consists of a masking layer, with mask value 0, followed by an LSTM with 256 hidden units. Its output then goes through a dropout layer, with rate 0.2, and a dense layer with 20 classes and a linear activation function.

For our MultiModal approach instead, a different 3D-CNN network is devised for each modality, and their outputs are then concatenated and inputted to the RNN part of the network.

The output of the network is then computed as a reduced mean of the output of the RNN.

\subsection{Loss Function and Evaluation Metrics}
The loss function used during training is an average logit loss one. It was implemented as a reduced mean of the sparse softmax cross entropy between the predictions and the labels.

As evaluation metric, the prediction accuracy was used. This is the number of correctly classified video clips over the entire test set which was predicted.

\subsection{Training}
The dataset was divided into training, validation and testing using a split division of 50/15/15 different splits. The model was trained using batches of 8 samples per iteration, for a total of 100 epochs. 

The training occurred on a cluster computer, for a maximum of 10 hours using 4 cores on a GeForce GTX 1080 Ti graphics card.

During training, an Adam optimizer is used to adapt the learning rate, which started as 1e-3, with an average logit loss over the validation accuracy. Additionally, to further optimize our model and prevent overfitting, we used early stopping, to terminate the training if there is no improvement after 20 epochs, and LR-Reduction, reducing the learning rate by a factor of 0.1 if the learning stagnates.

\section{Results}

\begin{table}[t]
\begin{tabular}{ |p{3cm}|p{2cm}|p{2cm}|  }
 \hline
 \textbf{Model} & \textbf{Data Aug.} & \textbf{Accuracy}\\
 \hline
 Skeleton & No & 0.522539\\
 MultiModalConv3DRNN & No &  0.7451701\\
 MultiModalConv3DRNN & Yes &  0.7644894\\
 Conv3DRNN-elu & No &  0.7635694\\
 \rowcolor{Gray}
 Conv3DRNN-swish & No &  0.772769\\
 \hline
\end{tabular}
\caption{Results of experiments}
\label{tab:res}
\end{table}
In Table \ref{tab:res}, the most relevant versions of our models are listed and evaluated.
Our Conv3DRNN with multiple modalities (RGB, skeleton and segmentation) with data augmentation performs better than the same model trained only on raw data. This was expected, since the model trained on raw data was overfitting, and with some data augmentation methods (adding Gaussian noise in this particular case) helped preventing that.
On the other hand, it was surprising to see that the single-mode Conv3DRNN with ELU on raw data performed the same as the MultiModal version of the model with data augmentation. The performance of single-mode Conv3DRNN then benefited from the use of a Swish activation function, in place of ELU, and this represents our best performing model.

\section{Conclusion and Future Work}
Dynamic gesture recognition is an important task to help humans with disabilities to better communicate with their surroundings. Our RGB only approach goes in the right direction looking at real-world use, since there aren't many cameras around which capture multimodal information.

Future work would be to go into detail, why data augmentation doesn't quite work on our approach and how we could fix that and the enhancement of the accuracy score with for example ensemble methods.

\bibliographystyle{IEEEtran}
\bibliography{howto-paper}

\begin{thebibliography}{1}
\providecommand{\url}[1]{#1}
\csname url@samestyle\endcsname
\providecommand{\newblock}{\relax}
\providecommand{\bibinfo}[2]{#2}
\providecommand{\BIBentrySTDinterwordspacing}{\spaceskip=0pt\relax}
\providecommand{\BIBentryALTinterwordstretchfactor}{4}
\providecommand{\BIBentryALTinterwordspacing}{\spaceskip=\fontdimen2\font plus
\BIBentryALTinterwordstretchfactor\fontdimen3\font minus
  \fontdimen4\font\relax}
\providecommand{\BIBforeignlanguage}[2]{{%
\expandafter\ifx\csname l@#1\endcsname\relax
\typeout{** WARNING: IEEEtran.bst: No hyphenation pattern has been}%
\typeout{** loaded for the language `#1'. Using the pattern for}%
\typeout{** the default language instead.}%
\else
\language=\csname l@#1\endcsname
\fi
#2}}
\providecommand{\BIBdecl}{\relax}
\BIBdecl

\bibitem{cnn}
Y.~Lecun, P.~Haffner, and Y.~Bengio, ``Object recognition with gradient-based
  learning,'' 08 2000.

\bibitem{rnn}
\BIBentryALTinterwordspacing
A.~Sherstinsky, ``Fundamentals of recurrent neural network {(RNN)} and long
  short-term memory {(LSTM)} network,'' \emph{CoRR}, vol. abs/1808.03314, 2018.
  [Online]. Available: \url{http://arxiv.org/abs/1808.03314}
\BIBentrySTDinterwordspacing

\bibitem{3dcnn}
S.~{Ji}, W.~{Xu}, M.~{Yang}, and K.~{Yu}, ``3d convolutional neural networks
  for human action recognition,'' \emph{IEEE Transactions on Pattern Analysis
  and Machine Intelligence}, vol.~35, no.~1, pp. 221--231, 2013.

\bibitem{ChaLearn}
\BIBentryALTinterwordspacing
Multimodal gesture recognition: Montalbano v2. [Online]. Available:
  \url{http://chalearnlap.cvc.uab.es/dataset/13/description/}
\BIBentrySTDinterwordspacing

\bibitem{mosaic}
\BIBentryALTinterwordspacing
S.~Escalera, J.~Gonz\`{a}lez, X.~Bar\'{o}, M.~Reyes, O.~Lopes, I.~Guyon,
  V.~Athitsos, and H.~Escalante, ``Multi-modal gesture recognition challenge
  2013: Dataset and results,'' in \emph{Proceedings of the 15th ACM on
  International Conference on Multimodal Interaction}, ser. ICMI ’13.\hskip
  1em plus 0.5em minus 0.4em\relax New York, NY, USA: Association for Computing
  Machinery, 2013, p. 445–452. [Online]. Available:
  \url{https://doi.org/10.1145/2522848.2532595}
\BIBentrySTDinterwordspacing

\bibitem{multimod}
\BIBentryALTinterwordspacing
C.~C. dos Santos, J.~L.~A. Samatelo, and R.~F. Vassallo, ``Dynamic gesture
  recognition by using cnns and star {RGB:} a temporal information
  condensation,'' \emph{CoRR}, vol. abs/1904.08505, 2019. [Online]. Available:
  \url{http://arxiv.org/abs/1904.08505}
\BIBentrySTDinterwordspacing

\bibitem{depthOnly}
\BIBentryALTinterwordspacing
K.~Lai and S.~N. Yanushkevich, ``Cnn+rnn depth and skeleton based dynamic hand
  gesture recognition,'' \emph{2018 24th International Conference on Pattern
  Recognition (ICPR)}, Aug 2018. [Online]. Available:
  \url{http://dx.doi.org/10.1109/ICPR.2018.8545718}
\BIBentrySTDinterwordspacing

\end{thebibliography}
\end{document}